\documentclass{article}
\usepackage{acra}
\usepackage{graphicx}
\usepackage{booktabs, multirow, hhline}
\usepackage[table]{xcolor}
\usepackage{amsmath}
\usepackage{tikz}
\usepackage{amssymb}
\usepackage{pifont}
 
\usepackage{amssymb}
\usepackage[ruled,vlined]{algorithm2e}
\usepackage{cite}
\usepackage{acra}
\usepackage{natbib}
\usepackage{hyperref}
\usepackage{url}

\title{VLM-Guided Visual Place Recognition for Planet-Scale Geo-Localization}

\author{
\small
Sania Waheed$^{1}$, Na Min An$^{2}$, Michael Milford$^{3}$, Sarvapali D. Ramchurn$^{1}$, Shoaib Ehsan$^{1,4}$ \\
$^{1}$Univ. of Southampton, UK \quad
$^{2}$KAIST, South Korea \quad
$^{3}$QUT, Australia \quad
$^{4}$Univ. of Essex, UK \\
\texttt{sw1m24@soton.ac.uk, naminan@kaist.ac.kr, michael.milford@qut.edu.au} \\
\texttt{sdr1@soton.ac.uk, s.ehsan@soton.ac.uk}
}

\begin{document}

\maketitle

\begin{abstract}
Geo-localization from a single image at planet scale (an advanced version of the kidnapped robot problem) is fundamental and challenging task in applications such as for navigation, autonomous driving, and disaster response due to vastly diverse locations, environmental conditions, and scene variations. Traditional retrieval-based geo-localization struggles with scalability and perceptual aliasing, while classification-based methods lack generalization and require extensive training data. Recent vision-language models (VLMs) offer a promising alternative by leveraging contextual understanding and reasoning. However, despite high accuracy, VLMs are prone to hallucinations and lack interpretability, making them unreliable as standalone solutions. We propose a novel hybrid geo-localization framework combining VLMs with retrieval-based visual place recognition (VPR) methods. Our approach uses a VLM to generate a prior that guides and constrains the retrieval search space. Next, we apply retrieval followed by a re-ranking step that selects geographically plausible matches based on feature similarity and proximity to the initially estimated coordinates. Evaluations on multiple geo-localization benchmarks show our method consistently outperforms prior state-of-the-art, particularly at street (up to 4.51\%) and city levels (up to 13.52\%). Our results demonstrate that VLM-generated geographic priors combined with VPR produce scalable, robust, and accurate geo-localization systems.
\end{abstract}

\section{Introduction}
\begin{figure} [tb]
    \centering
    \vspace*{1mm}
    \includegraphics[width=1\columnwidth]{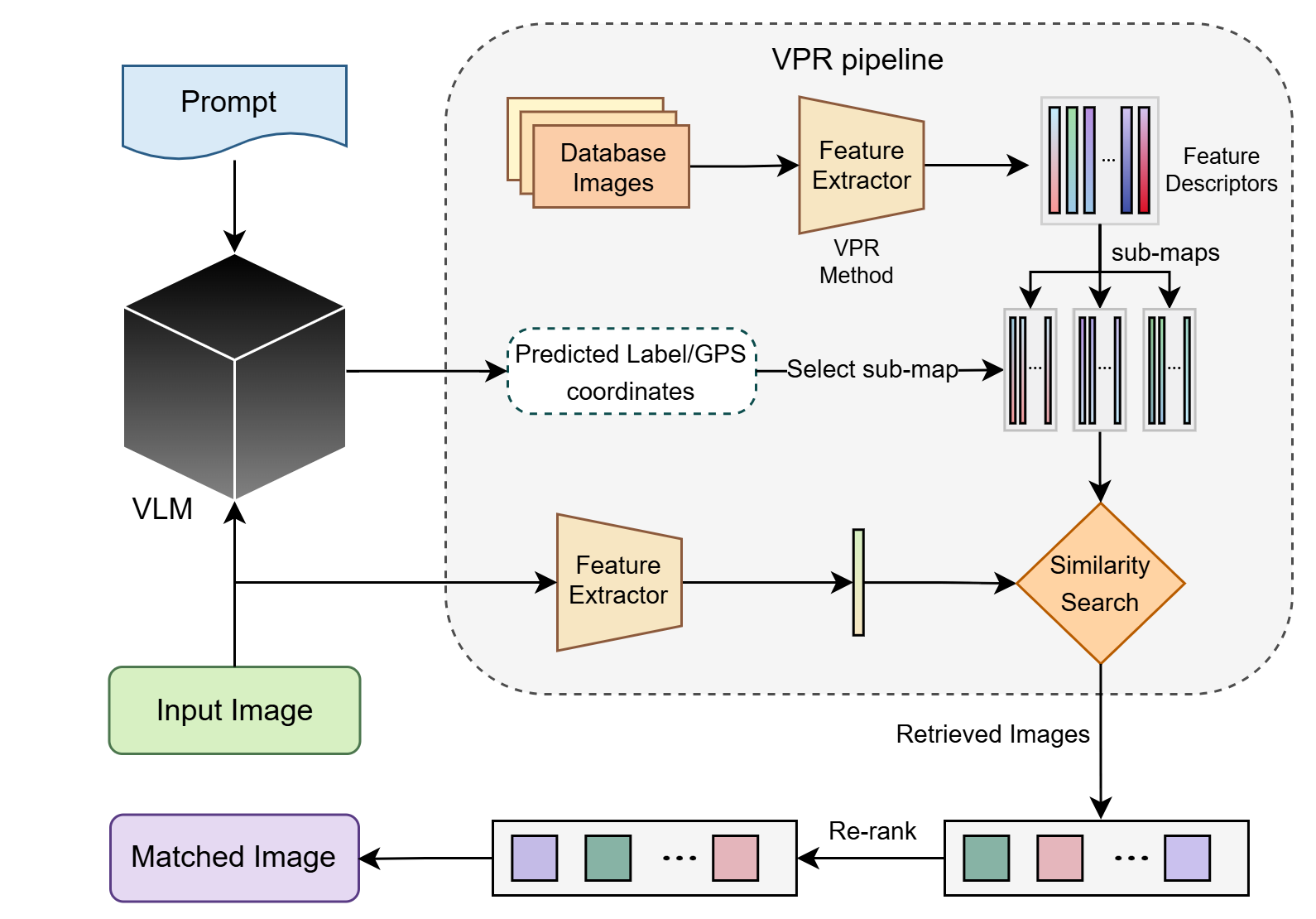}
    \caption{Block diagram of proposed pipeline for VLM-guided retrieval for image-based geo-localization. Feature descriptors for the reference images are first extracted using a VPR method and divided into sub-maps based on country or cluster-based partitions (details in Section \ref{sec:method}). A VLM predicts approximate GPS coordinates (VLM prior), which are used to select a relevant sub-map for retrieval. Descriptors for the query image are extracted using the same VPR method, and a similarity search is performed within the selected sub-map. Retrieved images are re-ranked based on their proximity to the predicted coordinates.}
    \label{fig:fig1-methodology}
\end{figure} 
Geo-localization is a fundamental yet challenging task in robotics applications such as navigation, autonomous driving, and search and rescue operations \cite{avola2024uav}. It is a generalized form of the \textit{kidnapped robot problem}, a long-standing robotics challenge where a robot is suddenly placed in an unknown location without any prior knowledge of its surroundings and must localize itself. The complexity of planet-scale image-based geo-localization arises from the vast diversity of locations, seasonal and environmental variations, and the visual ambiguity of many geographic regions. While distinct landmarks and unique landscapes provide strong location cues, many urban and rural areas exhibit high visual similarity, making precise localization difficult \cite{hays2008im2gps}.

Traditional geo-localization methods primarily fall into two categories: retrieval-based and classification-based approaches. Retrieval-based methods extract feature descriptors from a query image and match them against a large reference database to identify the most visually similar images. These methods generally perform well in landmark-rich environments \cite{suma2024amesasymmetricmemoryefficientsimilarity} but face significant challenges in visually ambiguous locations and suffer from scalability issues due to high database storage and search complexity \cite{vo2017revisiting}. Classification-based approaches, on the other hand, predict geographic coordinates by assigning the query image to a predefined geo-cell. However, they often struggle with fine-grained localization at street or city-level resolution and require extensive training data, leading to poor generalization \cite{muller2018geolocation, weyand2016planet, seo2018cplanet}.

More recently, Vision-Language Models (VLMs), ranging from CLIP \cite{radford2021learningtransferablevisualmodels} to GPT-4v \cite{achiam2023gpt}, have shown promising capabilities in geo-localization \cite{zhou2024img2loc, mendes2024granular, waheed2025image, haas2023learninggeneralizedzeroshotlearners, 10.1145/3557918.3565868, cepeda2023geoclipclipinspiredalignmentlocations}, offering contextual reasoning, environmental understanding, and broader geographic knowledge beyond visual similarity. However, VLMs alone remain unreliable as they make speculative or hallucinated guesses which are difficult to verify, have low interpretability, and can exhibit inconsistent behavior \cite{waheed2025image}. This motivates the need for a more structured approach that balances the contextual reasoning capabilities of VLMs with the robustness of retrieval-based methods.

To address these challenges, we propose a novel hybrid geo-localization method that combines VLM-generated predictions with retrieval-based refinement. Our approach first uses a VLM to generate an initial coordinate estimate, which serves as a strong prior to guide the retrieval process. A robust visual place recognition (VPR) method is then employed to find candidate matches within this constrained search space. Finally, we introduce a re-ranking mechanism that refines the retrieved results by selecting the most geographically plausible matches based on visual similarity and proximity to the VLM-generated prior. While VPR methods are not explicitly trained for geo-localization, they are designed to be robust to environmental variations, making them particularly well-suited for this task. However, their potential in geo-localization has been largely underexplored. We show that incorporating a strong prior from VLMs significantly improves their performance, achieving up to 4\% improvement on Im2GPS, 10.31\% on Im2GPS3k, and 2.6\% on GWS15k, making VPR methods practically viable for geo-localization. 

Our primary contributions in this paper are:
\begin{itemize}
    \item We propose a hybrid geo-localization framework that integrates VLM predicted coordinates as priors with VPR-based retrieval, combining contextual reasoning with robust visual matching.
    \item We introduce and evaluate strategies for creating constrained search spaces (``submaps'') guided by VLM priors, and demonstrate their impact across multiple VPR methods.
    \item We conduct extensive experiments to analyze the contribution of each component in our framework.
\end{itemize}

The remainder of this paper is organized as follows: Section~\ref{sec:related_works} reviews related work. Sections~\ref{sec:method} and \ref{sec:experiment} outline the proposed methodology and the experimental setup, respectively. Section~\ref{sec:results} presents the results and analysis, while Section~\ref{sec:conclusion} concludes the paper and discusses directions for future research.

\section{Related Works}\label{sec:related_works}
\subsection{Traditional Approaches to Geo-Localization}

Geo-localization from a single image remains a challenging problem, particularly at planet scale. Early work like \cite{hays2008im2gps} framed it as nearest-neighbor retrieval task using large-scale geotagged datasets. While retrieval-based methods perform well in constrained settings such as city-scale or landmark localization \cite{kendall2015posenet, wang2020online, weyand2020google, arandjelovic2016netvlad, avrithis2010retrieving, boiarov2019large}, they struggle to scale globally due to high computational costs and perceptual aliasing, where visually similar but geographically distant scenes lead to incorrect matches \cite{muller2018geolocation, regmi2019bridging}.
Classification-based methods \cite{muller2018geolocation, kordopatis2021leveraging, weyand2016planet, izbicki2020exploiting} reformulate the task as a multi-class classification problem, where the model predicts the geo-cell from a discrete set of cells that partition the Earth's surface. CNN-based models like \cite{weyand2016planet} demonstrated improved performance over retrieval methods by directly learning from discretized spatial partitions. However, this approach is sensitive to cell granularity: large cells reduce precision \cite{kordopatis2021leveraging}, while small ones struggle to learn discriminative features for each region due to class imbalance and visual ambiguity \cite{weyand2016planet, izbicki2020exploiting, zhou2024img2loc}. To mitigate this, hierarchical \cite{muller2018geolocation} and semantic partitioning strategies \cite{theiner2022interpretable} have been proposed, offering better spatial coherence and interpretability. Despite these advances, data imbalance remains a key challenge in classification-based geo-localization \cite{haas2024pigeonpredictingimagegeolocations}.

\subsection{Vision-Language Models (VLMs) for Geo-Localization}
VLMs have recently emerged as a promising direction in image-based geo-localization, offering semantic reasoning and contextual understanding that traditional vision-only methods lack \cite{zhou2024img2loc, jia2024g3, 10.1145/3557918.3565868, waheed2025image}. For example, \cite{zhou2024img2loc, jia2024g3} utilize retrieval-augmented generation (RAG) with GPT-4V \cite{achiam2023gpt} and LLaVA \cite{liu2023visual} to mitigate the hallucinated responses produced by VLMs. \cite{mendes2024granular, waheed2025image} show that GPT-4V outperforms fine-tuned VLM-based geo-localization models, suggesting that large-scale VLMs have an inherent knowledge of geographic distributions. However, their black-box nature, lack of interpretability, and possible hallucinations make them unreliable as standalone solutions. 

\subsection{Visual Place Recognition (VPR) Methods}
VPR is typically framed as an image retrieval problem \cite{arandjelović2016netvladcnnarchitectureweakly, 8099829, 9156836}, where a query image is localized by retrieving the closest match from a reference database. Among recent approaches, CosPlace \cite{berton2022rethinkingvisualgeolocalizationlargescale} employs a classification-based training strategy and uses the learned features for retrieval. MixVPR \cite{alibey2023mixvprfeaturemixingvisual} introduces a holistic aggregation technique that integrates global relationships into feature maps extracted from a pre-trained backbone. EigenPlaces \cite{berton2023eigenplacestrainingviewpointrobust} enhances the robustness of retrieval by improving viewpoint invariance through training on multiple views of the same location. BoQ \cite{alibey2024boqplaceworthbag} leverages a Transformer-based aggregation technique that learns global queries and applies cross-attention to probe local features from the backbone network. We choose these methods for our experiments as they provide complementary strengths in descriptor learning, robustness, and efficiency.

\section{Methodology}\label{sec:method}
This section presents the proposed hybrid geo-localization framework, which combines VLM predictions with VPR-based retrieval and re-ranking. As shown in Fig.\ref{fig:fig1-methodology}, a VLM generates an approximate geographic coordinate for a given query image. 

This prior constrains the retrieval search space to a relevant subset of reference images from the database, referred to as a \textit{submap}, which is then used for image retrieval. The top retrieved candidates are based on visual similarity between query and reference images and then re-ranked on the basis of geographic proximity to the VLM prior. This hybrid approach leverages the semantic understanding and contextual associations captured by VLMs along with the robustness of VPR methods to enable accurate and scalable image-based geo-localization. 

\subsection{Reference Set Preparation}\label{sec:method_1}
\subsubsection{Submap Construction}
To enable scalable and efficient retrieval, the reference dataset is divided into smaller geographically coherent subsets, referred to as \textit{submaps} and denoted by $\mathcal{S}$. Two partitioning strategies are considered:

\begin{itemize}
    \item \textbf{Country-based submaps:} Each reference image is assigned a country label via reverse geocoding\footnote{\texttt{reverse\_geocoder} library is used: \url{https://github.com/thampiman/reverse-geocoder}}, resulting in one submap $\mathcal{S}_c$ per country $c$. This provides a coarse but interpretable division of the dataset.

    \item \textbf{Clustering-based submaps:} K-means clustering is applied to the geographic coordinates of all reference images, producing $K$ submaps. Unlike country-based partitioning, this method captures data-dependent geographic coherence without relying on political boundaries:
    \begin{equation}
        \mathcal{S}_1, \dots, \mathcal{S}_K = \text{K-Means}(\{(lat_j, lon_j)\}_{j=1}^N, K)
    \end{equation}
    where $\mathcal{S}_k$ denotes the $k$-th submap, and $N$ is the total number of reference images. For our experiments, we set $K=100$.
\end{itemize}

\subsubsection{Feature Extraction and Indexing}
A VPR model $\Phi(\cdot)$ is used to extract image descriptors. For each image $I_r$ in the reference set $\mathcal{D} = \{(I_r, lat_r, lon_r)\}_{r=1}^N$, a $d$-dimensional descriptor is computed:
\begin{equation}
    f_r = \Phi(I_r), \quad f_r \in \mathbb{R}^d
\end{equation}

All descriptors are stored using FAISS indices\footnote{\url{https://github.com/facebookresearch/faiss}.}, organized by submap.

\subsection{Query Processing}

\subsubsection{VLM-Based Prior Estimation}
Given a query image $I_q$ and prompt $p$ \footnote{We use the LTM prompt from \cite{mendes2024granular}, which we omit here for brevity.}, a VLM generates a predicted geographic coordinate, which is used as the prior:
\begin{equation}
    (\hat{lat}, \hat{lon}) = \text{VLM}(I_q, p)
\end{equation}
The predicted coordinates $(\hat{lat}, \hat{lon})$ are used to select the most relevant submaps generated in the previous stage (Section~\ref{sec:method_1}).

\subsubsection{Query Feature Extraction and Retrieval}\label{sec:method_2}
The VPR model $\Phi(\cdot)$ extracts a descriptor from the query image $I_q$:
\begin{equation}
    f_q = \Phi(I_q), \quad f_q \in \mathbb{R}^d
\end{equation}

Similarity between the query descriptor $f_q$ and reference descriptors $f_r$ in the selected submap $\mathcal{S}$ is computed using the L2-squared distance:
\begin{equation}
    s(f_q, f_r) = \|f_q - f_r\|_2^2, \quad \forall f_r \in \mathcal{S}
\end{equation}

The top-$p$ most similar reference images are retrieved based on the smallest distances:
\begin{equation}
    \{r_1, r_2, ...., r_p\} = \{\operatorname{argsort}_{r \in \mathcal{S}}  s(f_q, f_r) \}_{i=1}^{p} \\
\end{equation}

\subsection{Geographic Re-ranking}
To refine the initial retrieval results, we rerank the top-$p$ candidates based on their geographic proximity to the VLM-predicted coordinates. Let $\{(\Tilde{lat_i}, \Tilde{lon_i})\}_{i=1}^p$ denote the coordinates of the retrieved images. The geographic distance $d_{i}$ to the VLM estimate is computed using the haversine formula:
\begin{equation}
    d_i = \text{Haversine}\left((\hat{lat}, \hat{lon}), (\Tilde{lat_i}, \Tilde{lon_i})\right), \quad \forall i \in \{1, \dots, p\}
\end{equation}

The candidates are then sorted in ascending order of $d_i$, and the best match $I^*$ is:
\begin{equation}
    I^* = \arg\min_i d_i 
\end{equation}

This re-ranking step enables not only the selection of the database images with high visual similarity (Section~\ref{sec:method_2}) but also ensures geographic proximity to the prior.

\section{Experimental Setup}\label{sec:experiment}
We evaluate our approach on three standard geo-localization benchmarks: IM2GPS \cite{hays2008im2gps}, IM2GPS3k \cite{vo2017revisiting}, and GWS15k \cite{clark2023werelookingatquery}. IM2GPS consists of 237 manually selected images from the IM2GPS6M dataset \cite{hays2008im2gps}, while IM2GPS3k includes 3,000 randomly sampled images from the same source, making it a bit more challenging. The GWS15k dataset is constructed by sampling countries in proportion to their surface area, randomly selecting a city within each, and retrieving Google Street View images from within a 5~km radius of the city center. Since this dataset is not publicly available, we reproduce it following the instructions provided in \cite{clark2023werelookingatquery}. Compared to the other two benchmarks, GWS15k offers a more geographically balanced distribution and poses more challenging localization scenarios. For the reference set, we use the MediaEval 2016 (MP-16) dataset \cite{7849098}, a standard in geo-localization tasks, which consists of 4.1 million geo-tagged Flickr images from across the globe.

To assess that our method's effectiveness stems from the underlying approach rather than a specific VLM, we evaluate it using two recent state-of-the-art (SoTA) models: GPT-4o (\verb|gpt-4o-2024-05-13|) \cite{achiam2023gpt} and Gemini-1.5-Pro. Both models are prompted using the Least-to-Most (LTM) prompting strategy introduced in \cite{mendes2024granular}. We extract the predicted geographic coordinates from the model outputs using the regular expression \texttt{r'[-+]?\textbackslash d*\textbackslash .\textbackslash d+|\textbackslash d+'}.

For the retrieval component, we employ four SoTA visual place recognition (VPR) methods: CosPlace \cite{berton2022rethinkingvisualgeolocalizationlargescale}, MixVPR \cite{alibey2023mixvprfeaturemixingvisual}, EigenPlaces \cite{berton2023eigenplacestrainingviewpointrobust}, and BoQ \cite{alibey2024boqplaceworthbag}. All methods use a ResNet-50 backbone to ensure a fair comparison. To maintain scalability, we set the feature dimension to 512 for CosPlace, MixVPR, and EigenPlaces. For BoQ, we use a feature dimension of 16,384, as it is the smallest available configuration.

Following standard evaluation protocols \cite{vo2017revisiting, muller2018geolocation, theiner2022interpretable, pramanick2022world, cepeda2023geoclipclipinspiredalignmentlocations, haas2023learninggeneralizedzeroshotlearners, clark2023werelookingatquery, zhou2024img2loc, haas2024pigeonpredictingimagegeolocations}, we report geolocation accuracy at five spatial scales: street-level (1~km), city-level (25~km), region-level (200~km), country-level (750~km), and continent-level (2500~km).

\section{Results}\label{sec:results}
\newcommand{\tikzxmark}{\ding{55}} 

\begin{table*}[!ht]
\vspace*{2mm}
    \centering
    \renewcommand{\arraystretch}{1.2}
    {\fontsize{14}{16}\selectfont
    \resizebox{\textwidth}{!}{
    \begin{tabular}{c|l|c|c|ccccc|ccccc|ccccc}
        \toprule  
        \multicolumn{4}{c|}{METHOD} & \multicolumn{5}{c|}{IM2GPS3k} & \multicolumn{5}{c|}{IM2GPS} & \multicolumn{5}{c}{GWS15k} \\
        \cmidrule(lr){1-4} \cmidrule(lr){5-9} \cmidrule(lr){10-14} \cmidrule(lr){15-19}
        \textbf{VPR} & \textbf{VLM} & \textbf{Submap} & \textbf{Re-rank} 
        & 1 km & 25 km & 200 km & 750 km & 2500 km
        & 1 km & 25 km & 200 km & 750 km & 2500 km
        & 1 km & 25 km & 200 km & 750 km & 2500 km \\
        \midrule

        \multirow{12}{*}{CosPlace} & \tikzxmark & \tikzxmark & \tikzxmark & 7.14 & 15.22 & 17.89 & 23.90 & 39.93 & 10.97 & 26.16 & 30.80 & 34.17 & 47.67 & 0.0 & 0.04 & 0.52 & 3.42 & 13.82\\
        & Gemini-1.5-pro & \tikzxmark & \checkmark & 9.03 & 20.87 & 28.27 & 41.27 & 59.23 & 18.14 & 40.08 & 51.48 & 74.26 & 90.72 & 0.03 & 1.72 & 12.61 & 40.26 & 73.65\\
        & GPT-4v & \tikzxmark & \checkmark & 12.63 & 30.67 & 45.07 & 64.57 & 83.5 & 19.41 & 42.19 & 54.85 & 76.37 & 92.41 & 0.02 & 1.62 & 11.68 & 38.21 & 72.5\\

        & Gemini-1.5-pro & Country & \tikzxmark & 8.82 & 26.4 & 37.79 & 58.46 & 79.04 & 13.5 & 35.44 & 49.79 & 70.46 & 86.92 & 0.04 & 1.75 & 9.67 & 34.52 & 71.42 \\
        & Gemini-1.5-pro & Cluster & \tikzxmark & 13.46 & 38.68 & 54.26 & 71.32 & 84.41 & 22.78 & 49.79 & 65.82 & 81.43 & 91.56 & 0.27 & 6.8 & 29.12 & 64.04 & 86.89\\
        & GPT-4v & Country & \tikzxmark & 9.78 & 26.73 & 37.87 & 58.39 & 79.51 & 13.50 & 34.18 & 49.37 & 68.78 & 87.34 & 0.04 & 1.75 & 9.67 & 34.52 & 71.42\\
        & GPT-4v & Cluster & \tikzxmark & 16.92 & 43.98 & 59.19 & 73.54 & 85.85 & 24.47 & 51.9 & 68.35 & 83.12 & \underline{94.09} & 0.29 & 7.91 & 29.77 & 63.44 & 85.09 \\
        
        & Gemini-1.5-pro & Country & \checkmark & 8.90 & 28.16 & 43.53 & 68.31 & 84.19 & 12.24 & 37.55 & 51.90 & 77.22 & 91.14 & 0.04 & 2.94 & 20.08 & 57.05 & 86.09 \\
        & Gemini-1.5-pro & Cluster & \checkmark & 14.71 & 39.78 & 55.22 & 71.4 & 84.34 & 22.78 & 48.95 & 66.24 & 80.17 & 91.56 & 0.27 & 9.34 & 32.01 & 64.69 & 87.13\\
        & GPT-4v & Country & \checkmark & 10.49 & 29.57 & 45.15 & 71.08 & 85.79 & 12.23 & 37.55 & 54.43 & 80.16 & \underline{94.09} & 0.06 & 2.82 & 19.09 & 55.60 & 84.19\\
        & GPT-4v & Cluster & \checkmark & 18.42 & 45.41 & \textbf{59.96} & \underline{73.67} & 85.95 & 24.89 & 51.48 & 69.2 & \textbf{83.97} & \underline{94.09} & 0.29 & 9.89 & 32.21 & 63.86 & 85.24\\
        \midrule

        \multirow{12}{*}{MixVPR} & \tikzxmark & \tikzxmark & \tikzxmark & 7.98 & 16.83 & 19.54 & 25.79 & 39.99 & 12.71 & 27.54 & 30.50 & 34.74 & 47.88 & 0.01 & 0.15 & 0.94 & 5.26 & 19.19 \\
        & Gemini-1.5-pro & \tikzxmark & \checkmark & 10.03 & 23.1 & 30.37 & 42.97 & 59.93 & 18.57 & 40.93 & 55.27 & 73.84 & 87.76 & 0.05 & 2.05 & 14.86 & 44.81 & 77.97 \\
        & GPT-4v & \tikzxmark & \checkmark & 13.47 & 31.2 & 44.97 & 64.77 & 83.5 & 20.25 & 42.62 & 57.81 & 74.26 & 88.61 & 0.07 & 2.06 & 14.09 & 42.49 & 76.98 \\
        
        & Gemini-1.5-pro & Country & \tikzxmark & 9.26 & 26.4 & 36.32 & 58.09 & 77.57 & 16.88 & 40.93 & 55.7 & 75.53 & 88.61 & 0.12 & 1.5 & 8.83 & 33.97 & 74\\
        & Gemini-1.5-pro & Cluster & \tikzxmark & 10.29 & 28.60 & 42.43 & 68.68 & 83.53 & 16.03 & 38.82 & 52.74 & 78.48 & 91.14 & 0.05 & 3.06 & 19.84 & 56.95 & 86\\
        & GPT-4v & Country & \tikzxmark & 10.61 & 27.46 & 38.47 & 59.33 & 78.88 & 17.72 & 40.51 & 54.43 & 73.84 & 89.03 & 0.12 & 1.57 & 9.36 & 34.70 & 73.69\\
        & GPT-4v & Cluster & \tikzxmark & 11.71 & 30.20 & 45.61 & 71.20 & 85.89 & 16.03 & 39.66 & 54.43 & 81.01 & \underline{94.09} & 0.08 & 2.38 & 18.29 & 54.99 & 84.22\\
        
        & Gemini-1.5-pro & Country & \checkmark & 14.19 & 38.31 & 54.63 & 71.4 & 84.41 & 22.36 & 48.52 & 67.09 & 80.59 & 91.56 & 0.35 & 6.97 & 29.56 & 64.48 & 86.99\\
        & Gemini-1.5-pro & Cluster & \checkmark & 14.63 & 40.22 & 55.07 & 71.4 & 84.34 & 21.1 & 49.37 & 67.51 & 81.01 & 91.56 & 0.34 & 9.41 & 31.98 & \underline{65.03} & \underline{87.15}\\
        & GPT-4v & Country & \checkmark & 17.88 & 43.84 & 59.09 & 73.51 & 85.95 & 24.89 & 51.05 & 69.62 & 83.12 & \underline{94.09} & 0.41 & 8.05 & 30.74 & 63.84 & 85.2\\
        & GPT-4v & Cluster & \checkmark & \textbf{19.25} & 45.41 & 59.69 & 73.54 & 85.99 & \textbf{26.16} & 53.16 & \underline{70.04} & \underline{83.54} & \underline{94.09} & 0.44 & 9.98 & \textbf{32.7} & 64.02 & 85.27\\
        \midrule
        
        \multirow{12}{*}{EigenPlaces} & \tikzxmark & \tikzxmark & \tikzxmark & 8.33 & 18.2 & 21.16 & 27.4 & 41.66 & 15.18 & 30.37 & 32.91 & 41.77 & 58.22 & 0.07 & 0.27 & 1.72 & 7.85 & 24.16 \\
        & Gemini-1.5-pro & \tikzxmark & \checkmark & 9.8 & 23.77 & 31.63 & 43.97 & 60.43 & 18.99 & 41.77 & 56.54 & 75.11 & 90.72 & 0.13 & 3.31 & 18.91 & 49.01 & 79.46 \\
        & GPT-4v & \tikzxmark & \checkmark & 14 & 33.17 & 46.87 & 65.37 & 83.67 & 20.25 & 42.62 & 58.23 & 75.95 & 91.98 & 0.14 & 3.19 & 17.93 & 46.95 & 78.29 \\
        
        & Gemini-1.5-pro & Country & \tikzxmark & 10.15 & 28.68 & 40.15 & 58.82 & 78.46 & 18.14 & 42.19 & 56.12 & 70.89 & 89.87 & 0.16 & 2.3 & 10.85 & 37.13 & 76.2 \\
        & Gemini-1.5-pro & Cluster & \tikzxmark & 10.37 & 29.78 & 44.26 & 68.46 & 84.26 & 18.14 & 39.66 & 53.16 & 78.06 & 91.56 & 0.01 & 2.76 & 19.28 & 56.92 & 85.84 \\
        & GPT-4v & Country & \tikzxmark & 11.11 & 29.1 & 41.11 & 59.83 & 79.58 & 18.14 & 41.77 & 54.43 & 68.78 & 90.3 & 0.2 & 2.34 & 11.48 & 37.8 & 75.73 \\
        & GPT-4v & Cluster & \tikzxmark & 12.04 & 31.93 & 47.04 & 70.67 & \textbf{86.15} & 18.14 & 40.5 & 55.27 & 80.59 & \underline{94.09} & 0.17 & 3.31 & 20.06 & 56.51 & 84.35 \\
        
        & Gemini-1.5-pro & Country & \checkmark & 13.75 & 38.75 & 54.78 & 71.25 & 84.41 & 20.68 & 49.37 & 65.82 & 80.17 & 91.56 & 0.47 & 7.77 & 30.65 & 64.81 & 87 \\
        & Gemini-1.5-pro & Cluster & \checkmark & 14.26 & 40.15 & 55.44 & 71.4 & 84.34 & 20.68 & 49.79 & 67.09 & 80.17 & 91.56 & 0.39 & 9.23 & 31.96 & 64.77 & \underline{87.15} \\
        & GPT-4v & Country & \checkmark & 17.48 & 44.38 & 59.03 & \textbf{73.71} & 85.85 & 23.21 & 53.16 & 69.62 & \underline{83.54} & \underline{94.09} & \underline{0.51} & 8.65 & 31.32 & 63.99 & 85.23 \\
        & GPT-4v & Cluster & \checkmark & 18.62 & \underline{45.65} & \underline{59.79} & \textbf{73.71} & 85.95 & 24.89 & 52.32 & \textbf{70.46} & \underline{83.54} & 93.67 & 0.48 & \textbf{10.28} & 32.61 & 64.14 & 85.25 \\
        \midrule

        \multirow{12}{*}{BoQ} & \tikzxmark & \tikzxmark & \tikzxmark &\textendash & \textendash & \textendash & \textendash & \textendash & \textendash & \textendash & \textendash & \textendash & \textendash & \textendash & \textendash & \textendash & \textendash & \textendash\\
        & Gemini-1.5-pro & \tikzxmark & \checkmark & \textendash & \textendash & \textendash & \textendash & \textendash & \textendash & \textendash & \textendash & \textendash & \textendash & \textendash & \textendash & \textendash & \textendash & \textendash\\
        & GPT-4v & \tikzxmark & \checkmark & \textendash & \textendash & \textendash & \textendash & \textendash & \textendash & \textendash & \textendash & \textendash & \textendash & \textendash & \textendash & \textendash & \textendash & \textendash\\
        
        & Gemini-1.5-pro & Country & \tikzxmark & 13.0 & 31.73 & 43.23 & 61.47 & 79.8 & 21.1 & 46.84 & 57.81 & 74.68 & 90.72 & 0.13 & 1.73 & 9.54 & 35.3 & 74.3 \\
        & Gemini-1.5-pro & Cluster & \tikzxmark & 13.31 & 33.8 & 48.28 & 70.97 & 86.02 & 21.52 & 44.73 & 59.07 & 82.28 & \underline{94.09} & 0.16 & 2.95 & 19.57 & 57.14 & 86.36 \\
        & GPT-4v & Country & \tikzxmark & 13.03 & 31.77 & 43.30 & 61.53 & 79.87 & 21.10 & 46.84 & 57.81 & 74.68 & 90.72 & 0.15 & 1.78 & 9.92 & 36.11 & 73.97 \\
        & GPT-4v & Cluster & \tikzxmark & 13.35 & 33.83 & 48.31 & 71.00 & \underline{86.05} & 21.52 & 44.73	& 59.07	& 82.28	& \underline{94.09} & 0.17 & 2.88 & 18.77 & 55.75 & 84.36 \\
        
        & Gemini-1.5-pro & Country & \checkmark & 14.03 & 34.93 & 47.9 & 66.07 & 82.0 & 21.94 & 50.63 & 65.82 & 80.59 & 92.41 & 0.45 & 7.39 & 30.19 & 64.7 & 86.97 \\
        & Gemini-1.5-pro & Cluster & \checkmark & 14.18 & 35.3 & 50.68 & 72.24 & \textbf{86.15} & 21.52 & 48.1 & 65.4 & \underline{83.54} & \textbf{94.51} & \textbf{0.52} & 9.56 & 32.27 & \textbf{65.17} & \textbf{87.21} \\
        & GPT-4v & Country & \checkmark & 18.33 & 45.27 & 58.97 & 73.6 & 85.83 & \underline{25.74} & \textbf{54.01} & \underline{70.04} & \underline{83.54} & \underline{94.09} & 0.45 & 8.41 & 31.05 & 63.78 & 85.21\\
        & GPT-4v & Cluster & \checkmark & \underline{18.92} & \textbf{46.11} & \underline{59.79} & \textbf{73.71} & 85.99 & 24.47 & \underline{53.59} & \textbf{70.46} & \textbf{83.97} & 93.67 & 0.46 & \underline{10.21} & \underline{32.62} & 64.19 & 85.28\\
        \bottomrule
    \end{tabular}
    }}
\caption{Performance comparison across three geo-localization benchmarks: IM2GPS3k, IM2GPS, and GWS15k. The table reports localization accuracy at multiple spatial resolutions, ranging from street-level (1~km) to continent-level (2500~km). Each row corresponds to a specific configuration of the proposed framework, combining a VLM (Gemini-1.5-pro or GPT-4v), a VPR method (CosPlace, MixVPR, EigenPlaces, BoQ), a submap strategy (none (\tikzxmark), Country-level, or Cluster-based), and whether re-ranking was done (\checkmark) or not (\tikzxmark). For BoQ, retrieval was only performed with submaps due to high feature dimensionality constraints. The highest value in each column is shown in \textbf{bold}, and the second-highest is \underline{underlined}.}
    \label{tab:modular_results}
\end{table*}

\begin{figure*}[h]
    \vspace{5mm} 
    \centering
    \includegraphics[scale=0.275]{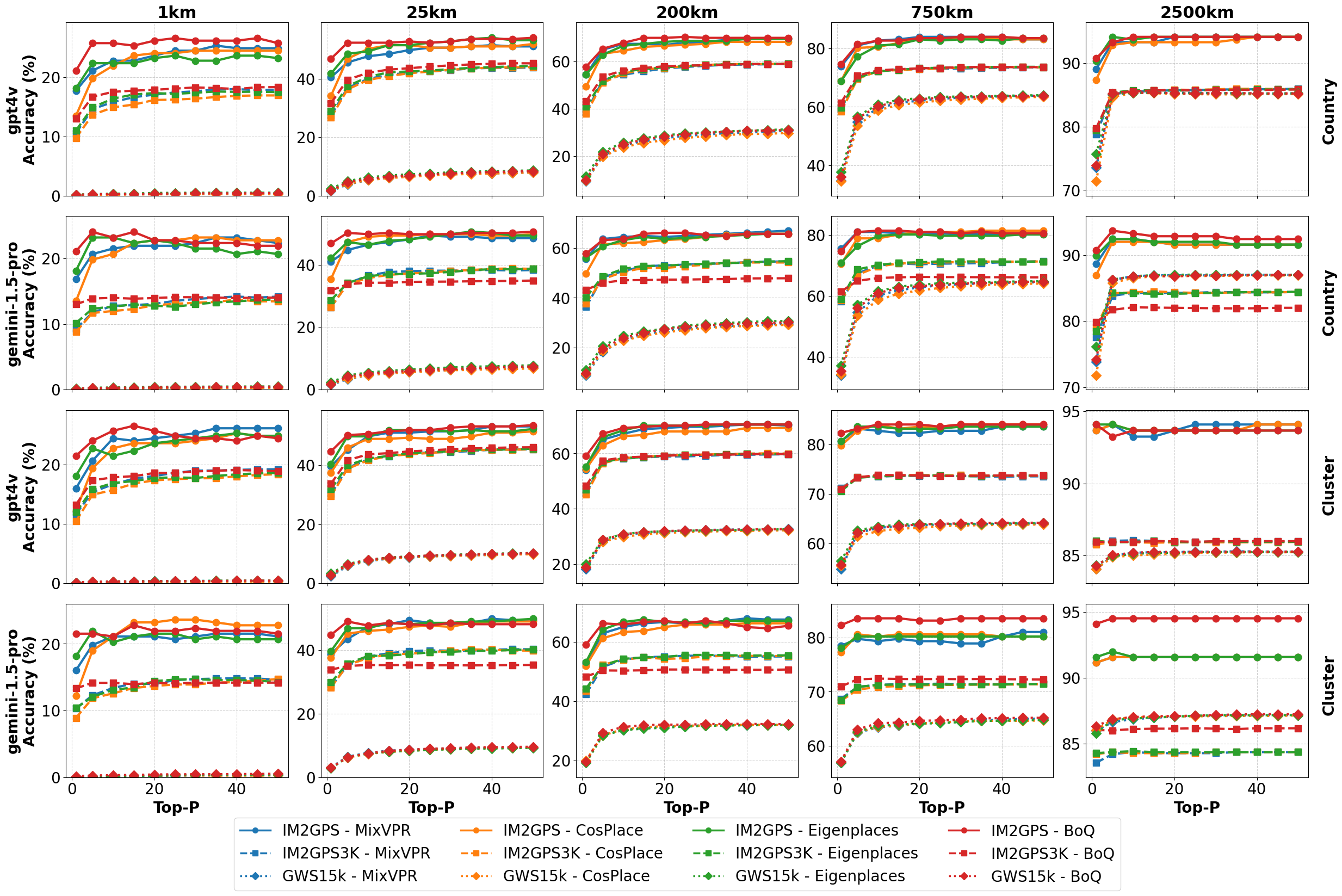}
    \caption{Top-p retrieval accuracy (\%) of four VPR methods, CosPlace, MixVPR, EigenPlaces, and BoQ, on benchmark datasets, IM2GPS, IM2GPS3k, and GWS15k, across multiple spatial resolutions (1~km to 2500~km). Retrieval is performed within submaps (country- and cluster-based) selected using VLM (GPT-4V and Gemini-1.5-Pro) prior. Accuracy improves with higher \textit{p} at finer spatial scales, but plateaus around \textit{p}=50. At coarser resolutions, increasing  \textit{p} has minimal effect. These trends are consistent across VLMs, submap types, and VPR methods.}
    \label{fig:top_p}
\end{figure*}

\begin{table*}[h]
    \centering
    \renewcommand{\arraystretch}{1.2}
    \resizebox{\textwidth}{!}{
    \begin{tabular}{l|ccccc|ccccc|ccccc}
        \toprule  
        \multirow{3}{*}{Method} & \multicolumn{5}{c|}{IM2GPS3k} & \multicolumn{5}{c|}{IM2GPS} & \multicolumn{5}{c}{GWS15k} \\
        & \textit{Street} & \textit{City} & \textit{Region} & \textit{Country} & \textit{Continent} & \textit{Street} & \textit{City} & \textit{Region} & \textit{Country} & \textit{Continent} & \textit{Street} & \textit{City} & \textit{Region} & \textit{Country} & \textit{Continent} \\
        & \textbf{1 km} & \textbf{25 km} & \textbf{200 km} & \textbf{750 km} & \textbf{2500 km} & \textbf{1 km} & \textbf{25 km} & \textbf{200 km} & \textbf{750 km} & \textbf{2500 km} & \textbf{1 km} & \textbf{25 km} & \textbf{200 km} & \textbf{750 km} & \textbf{2500 km} \\
        \midrule
        PlaNet & 8.5 & 24.8 & 34.3 & 48.4 & 64.6 & 8.4 & 24.5 & 37.6 & 53.6 & 71.3 & \textendash & \textendash & \textendash & \textendash & \textendash\\	  
        CPlanet & 10.2 & 26.5 & 34.6 & 48.6 & 64.6 & 16.5 & 37.1 & 46.4 & 62.0 & 78.5 & \textendash & \textendash & \textendash & \textendash & \textendash\\
        ISN & 10.5 & 28.0 & 36.6 & 49.7 & 66.0 & 16.9 & 43.0 & 51.9 & 66.7 & 80.2 & 0.05 & 0.6 & 4.2 & 15.5 & 38.5 \\	
        Translocator & 11.8 & 31.1 & 46.7 & 58.9 & 80.1 & 19.9 & 48.1 & 64.6 & 75.6 & 86.7 & 0.5 & 1.1 & 8 & 25.5 & 48.3\\
        GeoDecoder & 12.8 & 33.5 & 45.9 & 61.0 & 76.1 & 22.1 & 50.2 & 69.0 & 80.0 & 89.1 & \textbf{0.7} & 1.5 & 8.7 & 26.9 & 50.57\\ 	
        ZERO-SHOT STREETCLIP & \textendash & 22.4 & 37.4 & 61.3 & 80.4 & \textendash & 28.3 & 45.1 & 74.7 & 88.2 & \textendash & \textendash & \textendash & \textendash & \textendash \\	
        GeoCLIP & 14.11 & 34.47 & 50.65 & 69.67 & 83.82 & \textendash & \textendash & \textendash & \textendash & \textendash & 0.6 & 3.1 & 16.9 & 45.7 & 74.1 \\
        PIGETTO & 10.9 & 35.8 & 52.4 & 70.7 & 84.4 & 11.8 & 38.8 & 63.7 & 80.6 & 91.1 & \textbf{0.7} & 9.2 & 31.2 & \textbf{65.7} & 85.1\\ 
        
         Ours & \textbf{18.62} & \textbf{45.65} & \textbf{59.79} & \textbf{73.71} & \textbf{85.95} & \textbf{24.89} & \textbf{52.32} & \textbf{70.46} & \textbf{83.54} & \textbf{93.67} & 0.48 & \textbf{10.28} & \textbf{32.61} & 64.14 & \textbf{85.25} \\
        \bottomrule
        \end{tabular}
        }

    \caption{ Comparison with prior geo-localization methods on IM2GPS3k, IM2GPS, and GWS15k. Accuracy is reported at five spatial resolutions (1~km to 2500~km). Our method (EigenPlaces + Cluster-based submaps + re-ranking with GPT-4v) achieves state-of-the-art results across all datasets. Prior methods include PlaNet~\cite{weyand2016planet}, CPlaNet~\cite{seo2018cplanet}, ISN~\cite{muller2018geolocation}, TransLocator~\cite{pramanick2022world}, GeoDecoder~\cite{clark2023werelookingatquery}, Zero-shot StreetCLIP~\cite{haas2023learninggeneralizedzeroshotlearners}, GeoCLIP~\cite{cepeda2023geoclipclipinspiredalignmentlocations}, and Pigeotto~\cite{haas2024pigeonpredictingimagegeolocations}.
}
    \label{tab:comaprison_results}
\end{table*}

\begin{figure*}[h]
    \vspace{5mm} 
    \centering
    \includegraphics[scale=0.365]{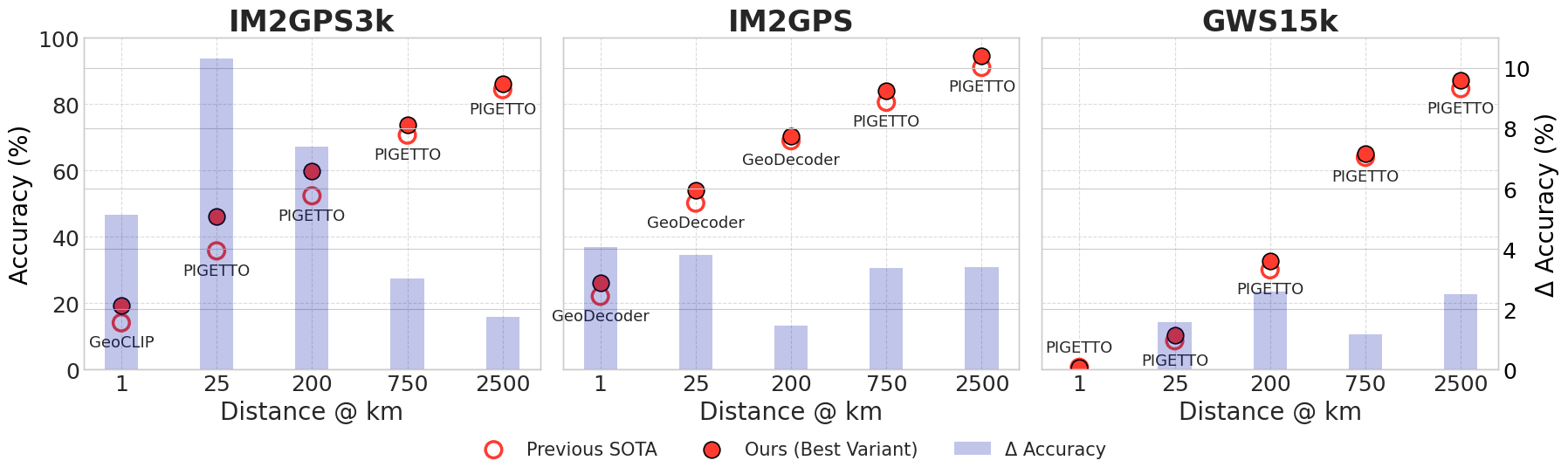}
    \caption{Accuracy comparison between the previous SoTA methods and our best variant across three geo-localization benchmarks: IM2GPS3k, IM2GPS, and GWS15k. Our approach consistently outperforms prior SoTA methods across all three datasets.}
    \label{fig:perf_comparison}
\end{figure*}

\begin{figure*}[tb]
    \vspace{5mm} 
    \centering
    \includegraphics[scale=0.19]{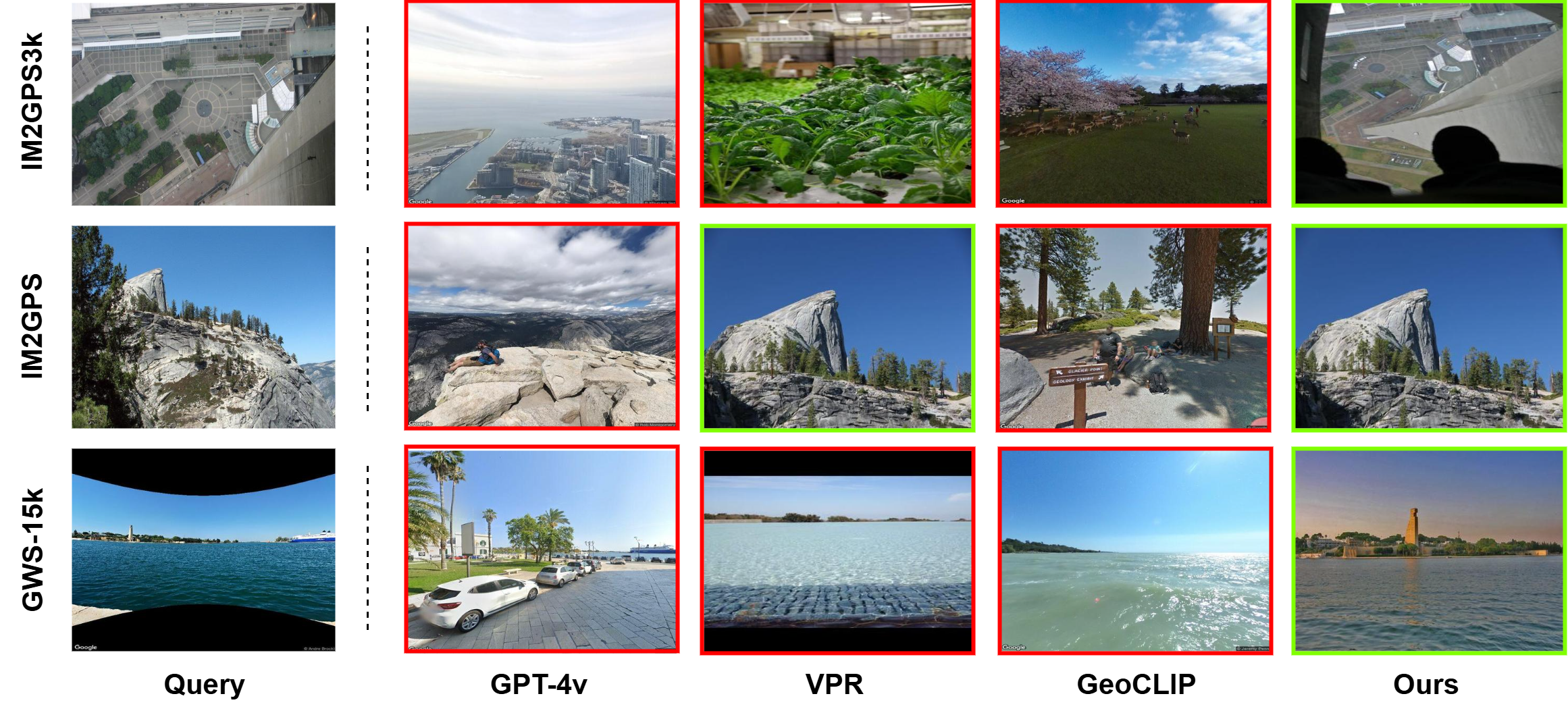}
    \caption{Qualitative comparison of geo-localization predictions. For each query image (leftmost column), we show the top-1 retrieved result from: (i) GPT-4v, (ii) VPR Method (EigenPlaces), (iii) one of the SoTA geo-localization methods: GeoCLIP, and (iv) our proposed approach (rightmost column). While VLMs and GeoCLIP directly predict coordinates, and VPR methods retrieve visually similar images, our method emphasizes spatially-aware retrieval with high visual overlap, making results easier to verify and more reliable. Correct predictions are highlighted in green, incorrect ones in red.}
    \label{fig:qualitative_comparison}
\end{figure*}
In this section, we compare our method against existing baselines and present a comprehensive ablation study to evaluate the contribution of each component in our framework.

\subsection{Comparison Across VPR Methods}
We compare four VPR methods within our framework: CosPlace, MixVPR, EigenPlaces, and BoQ. Table~\ref{tab:modular_results} shows that across all configurations, EigenPlaces and BoQ generally have stronger performance, particularly at finer resolutions, although CosPlace and MixVPR remain competitive as well. For instance, on IM2GPS3k, with cluster submaps and re-ranking, BoQ achieves 46.11\% accuracy at 25~km, slightly outperforming MixVPR (45.41\%), CosPlace (45.41\%), and EigenPlaces (45.65\%). The relative performance among VPR methods remains consistent across datasets, and the marginal differences between them are small compared to the larger gains obtained from our submap and re-ranking framework. This demonstrates that our pipeline is largely robust to the choice of VPR method, allowing the selection of a model based on resource or efficiency constraints without a major decrease in accuracy.

\subsection{Influence of the choice of VLM}
We evaluate the performance of two VLMs, GPT-4v and Gemini-1.5-Pro, in our framework. As shown in Table~\ref{tab:modular_results}, while GPT-4v generally performs slightly better on average, Gemini-1.5-Pro performs comparably and occasionally surpasses GPT-4v depending on the dataset and VPR pairing. Importantly, the framework exhibits similar performance trends using both VLMs, indicating that the architectural components (submaps, re-ranking) drive the majority of performance gains, rather than the specific VLM itself. This suggests that our method is VLM-agnostic, provided that the selected model offers reasonable spatial reasoning and world knowledge.

\subsection{Impact of Submaps}
Table~\ref{tab:modular_results} shows the impact of incorporating submap-based retrieval. Compared to global retrieval over the entire reference set, using submaps significantly improves localization accuracy across all datasets and distance thresholds, while also reducing retrieval time and computational cost. For instance, on the IM2GPS3k dataset, applying CosPlace with submaps increases accuracy from 15.22\% to 43.98\% at the 25~km threshold, and from 17.89\% to 59.19\% at 200~km. On average, submap-based retrieval provides an accuracy improvement of over 28\%. 

Gains from submaps are especially prominent at coarser resolutions (200–2500~km), where visual ambiguity increases and contextual understanding becomes crucial. Between the two submap strategies, cluster-based submaps consistently outperform country-level submaps (Table~\ref{tab:modular_results}), regardless of the VPR or VLM pairing. This reflects the advantage of finer-grained, data-driven partitioning over coarse administrative boundaries. While VPR methods alone often struggle to resolve such ambiguities when retrieving from the full global-scale reference set, VLMs provide semantically informed priors that help constrain retrieval to more likely geographic regions. As a result, retrieving within submaps increases the chance of identifying relevant images, even when exact visual matches are absent. This shows that constraining the search space leads to more semantically coherent matches.

\subsection{Effect of Re-ranking}
The final re-ranking step reorders the top-$p$ retrieved candidates in a computationally lightweight step based on their geographic distance to the VLM-predicted prior. As shown in Fig.~\ref{fig:top_p}, re-ranking accuracy improves with increasing $p$, saturating at around $p = 50$, beyond which additional candidates offer negligible gains. This indicates that the most relevant matches are typically already present within the top retrieved images.

Re-ranking generally enhances localization accuracy across all configurations. For example, in the MixVPR + re-ranking setup on IM2GPS3k, re-ranking increases street-level (1~km) accuracy from 7.98\% to 13.47\%, and region-level (200~km) accuracy from 19.54\% to 44.97\%. When applied to global retrieval, it improves performance by approximately 21\%, addressing cases where the top visual matches are misleading due to perceptual aliasing. By incorporating VLM priors into the final ranking, we prioritize semantically and geographically relevant candidates that may not have the highest visual overlap with the query image. 

While submap-based retrieval without re-ranking already performs well, adding re-ranking yields an additional 7\% improvement on average. Interestingly, submap retrieval without re-ranking often matches the performance of global retrieval with re-ranking, suggesting that simply narrowing the search space can resolve many of the ambiguities. However, the best results are consistently achieved when both are combined, showing that submaps and re-ranking are complementary and essential components of our approach.

\subsection{Comparsion with SoTA}
We compare our best-performing configurations on the IM2GPS, IM2GPS3k, and GWS15k benchmarks with SoTA methods. Table~\ref{tab:comaprison_results} presents a variant of our pipeline based on EigenPlaces with cluster-based submaps, and re-ranking using GPT-4v, against existing approaches across all three datasets, demonstrating competitive results across spatial resolutions.

Fig.~\ref{fig:perf_comparison}, highlights the performance improvement between our best-performing variant and the strongest prior method in each case, our approach achieves state-of-the-art results at nearly all spatial resolutions. The most substantial gains are observed on IM2GPS3k, where we outperform prior work by +10.3\% at the city level (25~km) and +7.6\% at the region level (200~km). On IM2GPS, we observe improvements of +4.1\% at street level (1~km) and +3.8\% at city level. Even on GWS15k, the most geographically diverse and challenging dataset, our method surpasses prior approaches by up to 2.6\% at region level and 2.5\% at continent level (2500~km). The only exception is a marginal drop of 0.18\% at the 1~km resolution, where our method slightly underperforms the best existing model. These performance gains stem from the complementary strengths of the proposed components. The inherent contextual understanding of VLMs allows the system to semantically narrow down the search space, improving retrieval precision, particularly in visually ambiguous cases. At the same time, the robustness of VPR methods ensures reliable place recognition despite challenging environmental variations.

Fig.~\ref{fig:qualitative_comparison} shows a visual comparison of the retrieved results of GPT-4v, EigenPlaces, GeoCLIP, and ours. GPT-4v and GeoCLIP directly predict geographic coordinates, from which we retrieve the corresponding Google Street View images, and EigenPlaces retrieves an image from the MP-16 dataset. While coordinate-generation methods can yield accurate locations, they does not guarantee visual overlap with the query. In contrast, our method, combines coordinate prediction with spatially constrained retrieval to produce results that not only correspond to the correct location but also exhibit strong visual overlap with the query image.

Notably, these results are achieved without any task-specific training; our framework leverages off-the-shelf VLMs and VPR methods. This design choice improves generalization, simplifies deployment at scale, and mitigates issues arising from distribution shifts that often undermine previous localization methods. These findings also underscore the potential of VPR methods for geo-localization, which despite not being explicitly tailored for this task, perform remarkably well when the retrieval space is constrained.

\section{Conclusion}\label{sec:conclusion}
This paper introduces a scalable image-based geo-localization method by combining the contextual understanding capabilities of VLMs with the robustness and efficiency of retrieval-based VPR methods. Through extensive experiments, we demonstrate that the proposed approach is modular, VLM-agnostic, and compatible with a range of SoTA VPR methods, achieving substantial improvements over existing methods, particularly at fine-grained spatial resolutions. Our framework requires no task-specific training, making it adaptable to diverse environments offering a scalable and generalizable solution for planet-scale geo-localization.

\bibliography{publications}
\bibliographystyle{named}

\end{document}